%% file: lrec-coling2024-example.tex
\documentclass[10pt, a4paper]{article}

\usepackage{lrec-coling2024} 

\title{Wisdom of Instruction-Tuned Language Model Crowds. \\Exploring Model Label Variation}

\name{\\\bf {Flor Miriam} {Plaza-del-Arco}, Debora Nozza, \bf Dirk Hovy}
\address{MilaNLP, Bocconi University, Department of Computing Sciences, Milan, Italy 
  \\
  \texttt{\{flor.plaza, debora.nozza, dirk.hovy\}@unibocconi.it}
}

\usepackage{microtype}

\usepackage{inconsolata}

\usepackage{xspace,mfirstuc,tabulary}
\usepackage{booktabs}
\usepackage{multirow}
\usepackage{graphicx}
\usepackage{tcolorbox}
\usepackage{multicol}

\newcommand{\promptbox}[3]{%
  \begin{tcolorbox}[colback=#3!10,colframe=black!50,title=#1] #2
  \end{tcolorbox}%
}

\abstract{
Large Language Models (LLMs) exhibit remarkable text classification capabilities, excelling in zero- and few-shot learning (ZSL and FSL) scenarios. However, since they are trained on different datasets, performance varies widely across tasks between those models. Recent studies emphasize the importance of considering human label variation in data annotation. However, how this human label variation also applies to LLMs remains unexplored. Given this likely model specialization, we ask: \textit{Do aggregate LLM labels improve over individual models (as for human annotators)?}
We evaluate four recent instruction-tuned LLMs as ``annotators'' on five subjective tasks across four languages.  
We use ZSL and FSL setups and label aggregation from human annotation.
Aggregations are indeed substantially better than any individual model, benefiting from specialization in diverse tasks or languages. Surprisingly, FSL does not surpass ZSL, as it depends on the quality of the selected examples. However, there seems to be no good information-theoretical strategy to select those.
We find that no LLM method rivals even simple supervised models.
We also discuss the tradeoffs in accuracy, cost, and moral/ethical considerations between LLM and human annotation.
 \\ \newline }

\begin{document}

\maketitleabstract

\section{Introduction}

Large Language Models (LLMs) have revolutionized many aspects of Natural Language Processing (NLP). \newcite{brown2020language} showed that LLMs have few-shot (FSL) and even zero-shot learning (ZSL) capabilities in text classification due to their extensive pre-training. Subsequent iterations have further improved these capabilities. 
Those improvements have seemingly obviated one of the most time- and labor-intensive aspects of NLP: annotating enough data to train a supervised classification model.
Instead, we can use LLMs to directly predict the labels via prompting. Indeed, various papers tested this hypothesis and found good performance on various NLP tasks \cite{zhao2023survey,su2022selective,wei2022emergent,brown2020language,plaza-del-arco-etal-2023-respectful}.
However, upon closer inspection, these claims require some caveats: different models excel at different tasks, datasets, and formulations \cite{gilardi2023chatgpt,tornberg2023chatgpt}. 

What if the answer is not to wait for one model to rule them all, though, but to treat their variation as specializations we can exploit, similar to the disagreement among human annotators? Different annotators have different strengths (or levels of reliability), and recent work \cite{basile-etal-2021-need,plank-2022-problem} has suggested using this human label variation to our advantage. We test whether the same applies to LLMs if we treat them as ``annotators''.

\input{prompts}

We use four state-of-the-art open-source instruction-tuned LLMs to assess their capabilities as ``annotators'': Flan-T5 \cite{chung2022scaling}, Flan-UL2 \cite{tay2023ul2}, T0 \cite{sanh2022multitask} (alongside its multilingual variant, mT0 \cite{muennighoff2022crosslingual}), and Tk-Instruct \cite{sanh2022multitask}. (We use open models to mitigate potential concerns regarding data contamination during the evaluation process and to facilitate replication.)
We evaluate them across five subjective prediction tasks (age, gender, topic, sentiment prediction, and hate speech detection) in four distinct languages: English, French, German, and Spanish. We use ZSL and FSL prompt instructions, similar to the ones we would give to human annotators.
For FSL, we explore different entropy-based strategies to choose the seed examples.
We then aggregate the LLM answers into a single label and evaluate their performance.

\textbf{We find that different models indeed excel on some tasks or languages, but not on others}. Some models even specialize on certain labels in a given task, but perform poorly on the others. \textbf{Their behavior thus mimics human expertise in wisdom-of-the-crowd settings.}

We then aggregate the model answers into a single label for each example.
The simplest approach is \textit{majority voting}: use the label that most LLMs suggested (ties are split randomly). However, the majority can still be wrong.
Instead, we can use \textit{Bayesian models of annotation} \cite{passonneau-carpenter-2014-benefits, paun-etal-2018-comparing} to weigh the answers based on the inferred reliability of each annotator. This approach is similar to Bayesian classifier combination, but does not require gold labels to assign the scores. Instead, it is completely unsupervised. That distinction is crucial, as we want to work with unannotated data.

In most cases, the aggregated labels of either method outperform even the best individual LLM. On average, aggregated labels are 4.2 F1-points better than the average LLM. However, even the best-aggregated performance is still well below that of even simple supervised models trained on the same data, and substantially lower than Transformer-based supervised models (by over 10 F1 points on average).


\textbf{In sum, aggregating several ZSL-prompted LLMs is better than using a single LLM}. Surprisingly, FSL-prompting is too varied to consistently improve performance. \textbf{However, treating LLMs as annotators cannot rival using human annotators for fine-tuning or supervised learning}.
We also discuss what these results mean for the role of human annotation and supervised learning in NLP, with respect to performance, but also time, cost, bias, and ethics.

\paragraph{Contributions} 
(\textbf{1}) We explore the feasibility of four open-source instruction-tuned LLMs as ``annotators'' via ZSL and FSL prompting on five subjective classification tasks; 
(\textbf{2}) we compare them across four languages; 
(\textbf{3}) we analyze the robustness of two label aggregation methods to check whether we can benefit from model label variation in subjective tasks; and 
(\textbf{4}) we discuss the technical, moral, and ethical ramifications of this development for NLP and annotation.

\section{Data}
For our experiments, we use two datasets: \textbf{Trustpilot} \cite{hovy2015user} and \textbf{HatEval} \cite{basile-etal-2019-semeval}. Note that for most models, these datasets are ``unseen''
, i.e., the data was not part of the LLMs' training. The one exception is HatEval in EN, which is included in Flan-T5 and Flan-UL2 models. We aim to evaluate their performance in a data contamination scenario,  offering insights into models' generalization capabilities unaffected by such contamination.

\textbf{Trustpilot} \cite{hovy2015user} is a multilingual dataset with demographic user information from various countries. It uses reviews from the user review website Trustpilot. 
To test a variety of languages commonly found in LLMs, we select data with English from the United States, German from Germany, and French from France for our experiments.
The data includes labels for sentiment, the topic of the review, and two demographic dimensions of the review authors: self-declared gender and age (these two are not available for all data points). We use the same data splits as \newcite{hovy-2015-demographic} to ensure comparability and consistency. Given our ZSL setup, we evaluate on their evaluation sets for each language, which consists of the joint development and test sets. 

\textbf{HatEval} \cite{basile-etal-2019-semeval} is a multilingual dataset for HS detection against immigrants and women on Twitter, part of a SemEval 2019 shared task. The dataset contains Spanish and English tweets manually annotated via crowdsourcing. 
We use the benchmark test set provided by the HatEval competition for both languages.

\subsection{Tasks}
We evaluate the performance of LLMs as annotators on five prediction tasks: four from the Trustpilot dataset and one from the HatEval corpus. These tasks involve sentiment analysis, topic detection, and predicting demographic attributes (gender and age). These two \textbf{attribute classification (AC)} tasks are binary: the \textbf{gender} of the text author (\textit{male} or \textit{female})\footnote{The data does not allow a more fine-grained classification of gender identities, as the original website only provided users with those two options. See Ethical Considerations for more discussion.} and the \textbf{age} of the text author (\textit{under 35} or \textit{above 45} years old). In the \textbf{sentiment analysis (SA)} task, reviews are classified into \textit{negative}, \textit{neutral}, and \textit{positive} sentiments based on 1, 3, and 5-star ratings, respectively. The \textbf{topic detection (TD)} task uses the review categories of the texts to classify them into one of five topics. For this task, the exact topics vary across languages.

\textbf{EN}: \textit{Car lights}, \textit{fashion accessories}, \textit{pets}, \textit{domestic appliances}, and \textit{hotels}. \textbf{DE}: \textit{Wine}, \textit{car rental}, \textit{drugs} and \textit{pharmacy}, \textit{flowers}, and \textit{hotels}. \textbf{FR}: \textit{Clothes and fashion}, \textit{fashion accessories}, \textit{pets}, \textit{computer and accessories}, and \textit{food and beverage}.

For the \textbf{hate speech detection (HS)}, the task is to classify a tweet as either hate speech or non-hate speech.

\section{Models}

We experiment with four state-of-the-art instruction-tuned LLMs from the same model family, the T5 with an encoder-decoder \cite{raffel2020exploring} architecture. 
We specifically select these models because they were fine-tuned on a diverse range of instructions. They use intuitive explanations of the downstream task to respond to natural language prompts, similar to the instructions provided to human annotators. 
Furthermore, these models are all open-source, letting us inspect the training data and examine data contamination.
Our selection represents a realistic LLM annotator pool for a current NLP practitioner. As models evolve rapidly, though, this selection is likely to change. However, the results from using a diverse pool of LLM annotators should hold regardless.

In particular, we use the following models:

\begin{itemize}
    \item \textbf{Flan-T5} \cite{chung2022scaling} is a sequence-to-sequence transformer model built on the T5 architecture \citep{raffel2020exploring}. The model has been pre-trained with standard language modeling objectives and subsequent fine-tuning on the extensive FLAN collection \citep{longpre2023flan}. The FLAN collection contains more than 1,800 NLP tasks in over 60 languages. We use the largest version\footnote{{\url{https://huggingface.co/google/flan-t5-xxl}}} of this model. 
    \item \textbf{Flan-UL2}  \cite{tay2023ul2} is the Flan version of the T5 and UL2 model. It has a similar architecture to T5, but with an upgraded pre-training procedure known as UL2\footnote{\url{https://huggingface.co/google/flan-ul2}}. 
    \item \textbf{T0}   \cite{sanh2022multitask} and the multilingual \textbf{mT0} \cite{muennighoff-etal-2023-mteb}. T0\footnote{\url{https://huggingface.co/bigscience/T0}} is an encoder-decoder model based on T5 that is trained on a multi-task mixture of NLP datasets over different tasks. For the non-English languages, we use mT0\footnote{\url{https://huggingface.co/bigscience/mt0-xxl}} since T0 has been trained on English texts. mT0 is based on Google's mT5 \cite{xue-etal-2021-mt5} and has been fine-tuned on xP3\footnote{\url{https://huggingface.co/datasets/bigscience/xP3}}, which covers 13 training tasks across 46 languages with English prompts.
    \item \textbf{Tk-Instruct} \cite{wang-etal-2022-super} is a generative model for transforming task inputs given declarative in-context instructions, like ``\textit{Given an utterance and the past 3 utterances, output ‘Yes’ if the utterance contains the small-talk strategy, otherwise output ‘No’. Small-talk is a cooperative negotiation strategy...}'' \citep[adapted from][]{wang-etal-2022-super}. It is also based on T5 but trained on all task instructions in a multi-task setup. It is fine-tuned on the \texttt{SUPER-NATURALINSTRUCTIONS} dataset \cite{triantafillou2020metadataset}, a large benchmark of 1,616 NLP tasks and their natural language instructions. It covers 76 task types across 55 different languages\footnote{\url{https://huggingface.co/allenai/tk-instruct-3b-def}}.
\end{itemize}

\paragraph{Computing Infrastructure} We run all experiments on a server with three NVIDIA RTX A6000 and 48GB of RAM.
 

\subsection{Prompting}
A prompt is an input that directs an LLM's text generation, ranging from a single sentence to a paragraph. It guides the model's comprehension and influences its output. Figure \ref{fig:prompts} depicts the task formulations (prompt instructions) we give to the LLMs, who act as our annotators, for every considered text classification task. 
We add ``Answer'' to mark the output field after the instruction to improve the LLMs' understanding and output format. 
For the TD tasks, the list of five topics varies by language. For instance, the prompt for the English TD task is: ``I love the earrings I bought,'' ``Is this review about `car lights,' `fashion accessories,' `pets,' `domestic appliances,' or `hotels'?'' <\texttt{Answer}>: \{LM answer\}.

Tk-Instruct requires a prompt template with specific fields: ``definition'', ``input,'' and ``output.''
The ``definition'' is used to specify the instruction or guidance, the ``input'' contains the instance to be classified, and the ``output'' is the output indicator. 
For instance, the prompt for the HS task is the following: <\texttt{Definition}>
Is this tweet expressing ``hate speech'' or ``non-hate speech?'' <\texttt{Input}> ``I hate you''
<\texttt{Response}>: \{LM answer\}.

We use task-specific prompts to assess the model's performance on the resulting outputs for zero- and few-shot prediction. We used default parameters for the models.

If the output does not correspond to a valid class, we assign the most common class for that task.
For instance, these out-of-label (OOL) predictions vary between tasks and models for the ZSL setup.
Binary or ternary classification tasks (AC, SA, HS) exhibit a very low OOL percentage (<1\%). In contrast, TD shows a significantly higher percentage (13\%) due to the larger number of classes and their more descriptive nature (e.g., ``fashion accessories'').
At the model level, Flan models have a very low OOL percentage (1\%), T0 and Tk-Instruct have a low OOL percentage ($\sim$2\%).


\subsection{Baselines}

We compare the LLMs across our five tasks to two baselines: \textit{the most frequent class} and \textit{random choice}.

The \textit{most frequent class} baseline does not require any model. It always picks the most frequent label for a task as final prediction. This standard baseline method is very strong in unbalanced datasets. However, it requires knowledge of the label distribution.
The \textit{random-choice} method randomly picks a label from the set of labels for a task. It represents a lower bound. 

\subsection{Aggregation of Labels}
For each example, we get four labels: one from each LLM annotator.
We use two different methods to aggregate these four labels into a single label: \textit{majority voting}, and \textit{a Bayesian model of annotation}, Multi-Annotator Competence Estimation \citep[MACE,][]{hovy-etal-2013-learning}. These methods use different aggregation methods. Both are common in the literature \cite{klie-etal-2023-lessons}.

\textit{Majority-voting} selects the label returned most by the four models. In case of a tie, it randomly chooses among the top candidates. This approach is common in many annotation projects, but has the drawback that the majority can still be wrong.

\textit{MACE} is a Bayesian annotation tool that computes two scores: the competence (reliability) of each annotator (i.e., the probability an annotator chooses the ``true'' label based on their expertise instead of guessing one) and the most likely label.
MACE uses variational Bayesian inference to infer both variables, and works on unlabeled data. The aggregated MACE labels are usually more accurate downstream than majority voting, and competence scores correlate strongly with actual annotator proficiency \cite{paun-etal-2018-comparing}.
Competence scores tend to correlate with annotators' actual expertise \cite{hovy-etal-2013-learning}, and can therefore be used to directly compare annotator quality in the absence of gold labels.
As a probabilistic model, it also computes the entropy of each example, including both annotator competence and agreement. 
It is therefore a proxy for how ``difficult'' an example is to label. 
We use MACE to get the competence of each LLM and the entropy of each example, which we use to select seed examples for FSL.

\section{Results}


We compare the four models as annotators along several dimensions:

\textbf{How much do models agree with each other?} This assesses the consensus among them and indicates specialization.
\textbf{How reliable is each model?} This evaluates the consistency and trustworthiness of individual model predictions, key for label aggregation.
\textbf{How accurate are the predictions of the individual models versus their aggregations?}
This last question assesses the prediction quality of individual LLMs vis-a-vis aggregations to determine whether this approach is a viable alternative to supervised learning.

We first report ZSL results and then discuss the FSL setting separately (Section \ref{sec:fs_results}).

\subsection{Inter-model Agreement}
To assess the level of specialization among the LLMs as annotators, we evaluate their agreement. We use four common inter-annotator-agreement metrics: Cohen's $\kappa$ \cite{Cohen1960ACO}, Fleiss' $\kappa$ \cite{Fleiss1971MeasuringNS}, and Krippendorff's $\alpha$ \cite{krippendorff1computing} (which all correct the observed agreement for expected agreement), as well as the unweighted raw agreement (i.e., the uncorrected level of agreement between LLMs). The results are shown in Table \ref{tab:agreement_trustpilot}.
Note that the number of labels does not factor into agreement, and that raw agreement is usually higher than chance-corrected inter-annotator agreement measures.

\input{inter_model_agg_trustpilot}

The results show a wide range of agreement values, but a few takeaways become apparent: 

1) \textbf{The scores suggest that the different models specialize on different tasks and labels}. As we will see in the performance and reliability analysis, some models perform better on some tasks than others. Model specialization suggests that aggregation is likely beneficial (as the aggregation hopefully benefits from differing expertise).

2) \textbf{Language does not factor into the differences}. The models we test are all multi-lingual, and the languages we test are generally high-resource. The agreement difference between the different languages on the same task is negligible.

3)  \textbf{Some tasks show higher agreement than others}: SA has higher scores than TD, and the others have little to no agreement. However, we do not know whether high-agreement tasks are inherently easier, or whether the models are all wrong in the same direction. 

\input{competence_trustpilot}

\subsection{Reliability}
When aggregating specialized annotators, we might want to trust more specialized ones more.
We use the competence scores from MACE to assess the reliability of each model. 
Table \ref{tab:tab_competence_trustpilot} shows the competence scores.

The competence scores support the specialization hypothesis for the different models on different languages and tasks. \textbf{No model is dominant in all settings, though the Flan models tend to have higher competence scores than the other models} (reflected in their higher mean competence scores).

\input{zs_results}

\subsection{Model Performance and Robustness of Label Aggregation}
Ultimately, we care about the predictive performance of the annotator method. Table \ref{tab:zs_results} shows the macro-F1 scores of the LLMs on all tasks and languages.
We compute the statistical difference of the individual results over the random-choice baseline, using a bootstrap sampling test with the \textit{bootsa}\footnote{\url{https://github.com/fornaciari/boostsa}} Python package. We use 1,000 bootstrap samples, a sample size of 20\%, and $p \leq 0.01$. 

\textbf{Most models clearly and significantly outperform the random-choice and even most-frequent-label baselines}. 
Note though that Flan-T5 and Flan-UL2 included the HatEval dataset in their training. Consequently, they perform substantially better than the other models (with Flan-T5 receiving a very high competence score from MACE).

\paragraph{Aggregation}
When aggregating annotations into a single label, we implicitly assume that a) there is a single correct answer and b) the wisdom of the crowd will find it. The first assumption is up for debate \cite{basile-etal-2021-need}, but the latter is clearly borne out by the results here.
\textbf{On average and in most individual cases, majority voting and MACE aggregation predictions are better than most models}. In 6 out of 14 tasks, MACE was the best model.
Note that for SA, Tk-Instruct performs better than the aggregation methods in all languages. For AC-Gender in English, Flan-UL2 is better, and in German, no method outperforms random choice (though MACE aggregation is close).

Overall, the two aggregation methods are substantially more robust than any one individual model across all languages and datasets (see the Mean results in Table \ref{tab:zs_results}). Presumably, they suffer less from the variance across tasks and languages and instead are able to exploit the specialization of each model as a source of information.
The MACE competence score correlates with the actual performance of the models: 0.93 Spearman $\rho$ and 0.83 Pearson $\rho$. 
This correlation suggests that MACE identified the model specializations correctly. A custom weighting of each model's prediction (for example, based on the actual performance) might perform even better. In practice, though, this weighting would of course be unknown.

\paragraph{Comparison to supervised learning}
ZSL holds a lot of promise for quick predictions, but to assess its worth, we need to compare it to supervised models based on human annotation.
For the Trustpilot data, we compare our best ZSL result for each task and language (see Table \ref{tab:zs_results}) to two supervised models. 1) a simple Logistic Regression model \citep[the baseline ``agnostic'' results reported in][]{hovy-2015-demographic} and a recent Transformer-based model \citep[the best results from][]{hung-etal-2023-demographic}.
Similarly, for HatEval, we compare with 1) a simple linear Support Vector Machine based on a TF-IDF representation \citep[the baseline results reported in][]{basile-etal-2019-semeval} and a fine-tuned multilingual Transformer model \cite{nozza-2021-exposing}. Table \ref{tab:results_trustpilot_comp} shows the results. 
The two methods approximate an upper and lower bound on supervised learning for these datasets.
\input{results_comp_supervised}

Except for 4 cases (AC-Gender and AC-Age in French, AC-Gender in English, HS in English), even the simple ML models beat the best ZSL result we achieved, be it from an LLM or aggregation method. 
Compared to the upper bounds from \newcite{hung-etal-2023-demographic}, we see an average performance gap of 10.5 F1 points.
Only for HS in English ZSL achieves better results, likely attributed to the data contamination found in the Flan models. 

\textbf{These results show that while ZSL might be a fast approximation for prediction tasks, it is still far from competitive with supervised learning}.

\subsection{Few-shot Learning}\label{sec:fs_results}
FSL has the potential to perform better than ZSL, so we ask whether using FSL models as annotators improves over our ZSL experiments. 
We investigate whether providing a limited set of examples enhances annotation capabilities, similar to instructing human annotators. The short answer is no.
We apply this method to the English Trustpilot dataset.

To choose the seed examples, we compare three methods. Random selection and selection based on the MACE entropy scores\footnote{\url{https://github.com/dirkhovy/MACE}}. Entropy lets us identify two groups of examples: (1) maximum entropy indicates models were less confident or disagreed more, indicating higher difficulty, and (2) low entropy indicates models were more confident or agreed more, indicating lower difficulty. 
For each task and label, we randomly choose 4,000 instances\footnote{We use 4,000 instances to make things as comparable as possible.} and use MACE to compute the entropy for each instance. From the initial pool of 4,000, we sample three exemplars per class based on the method (low entropy, max entropy) and use these as few-shot seeds, prepending them to the prompt. We compare these results to the random baseline.

\begin{figure*}
    \centering    \includegraphics[width=\textwidth]{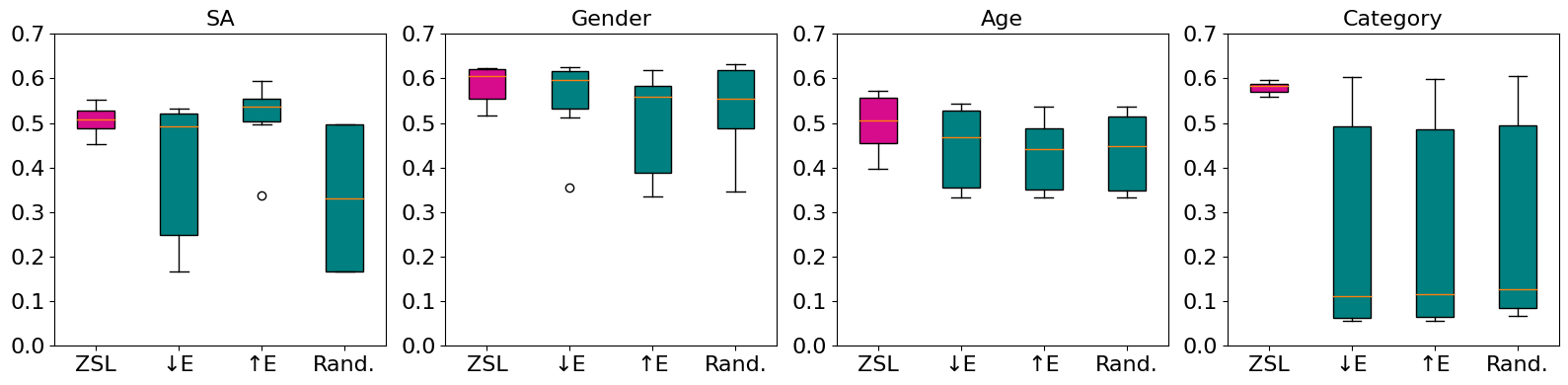}
    \caption{ZSL vs.\ FSL Macro-F1 scores on English Trustpilot tasks. FSL sample selection strategies: Low Entropy ($\downarrow$ E), Max Entropy ($\uparrow$ E), and Random (Rand). All FSL methods show much greater variance than ZSL.}
    \label{fig:zsl_vs_fsl}
\end{figure*}

Figure \ref{fig:zsl_vs_fsl} shows a comparison between the ZSL and FSL approaches. \textbf{Our analysis reveals that there are no statistically significant differences between the two prompting methods}. In general, though, FSL does not perform as well as ZSL across subjective tasks.
Within FSL, a prominent pattern emerges: it exhibits notably higher variance across tasks than ZSL. Presumably, exemplar quality heavily influences performance. 

Regarding the two entropy-based selection methods, our results show no consistent trends between them. 
The choice is somewhat task-dependent: Max entropy seems to perform well for SA and AC-Gender tasks, while low entropy works best for AC-Age and TD tasks. The random strategy is less consistent across tasks. This discrepancy further underlines the inherent challenge in selecting `good' exemplars for FSL. 
Our findings suggest that using no exemplars (ZSL) is generally more stable and consistent for aggregation.

\section{Related Work}
Generating human-annotated data is time-consuming and costly, especially for complex or specialized tasks with limited available data. Instead, a possible solution is leveraging automatic annotation models, often using a small subset of labeled data \cite{smit-etal-2020-combining,rosenthal-etal-2021-solid}, known as `weak supervision' \cite{stanford-ai-weak-supervision}. Supervised learning has emerged as the dominant method, driven by the widespread adoption of traditional machine learning models and Transformer-based models like BERT \cite{devlin-etal-2019-bert}.

More recently, LLMs have shown zero-shot and few-shot learning capabilities \cite{brown2020language}. Researchers have further advanced these models with natural language instructions \cite{chung2022scaling, wang-etal-2022-super,alpaca}, enabling innovative techniques like prompting \cite{liu2023pre} without the need to train a supervised model. Several papers explored these new techniques with promising performance on various NLP tasks \cite{brown2020language,plaza-del-arco-etal-2022-natural-emotions,su2022selective,sottana-etal-2023-evaluation}. 
Recent work has focused on exploring the capabilities of LLMs as annotators. For instance, \citet{lee2023large} evaluate the performance and alignment between LLMs and humans, revealing that these models not only fall short in performing natural language inference tasks compared to humans but also struggle to capture the distribution of human disagreements accurately. Other studies have used ChatGPT as an annotator. Some report excellent capabilities \cite{Huang_2023,gilardi2023chatgpt,tornberg2023chatgpt,he2024if}, but \newcite{kuzman2023chatgpt} found that ChatGPT's performance notably drops for less-resourced languages. Similarly, \citet{kristensenmclachlan2023chatbots} show that on two seemingly simple binary classification tasks, the performance of ChatGPT and open-source LLMs varies significantly and often unpredictably, and supervised models systematically outperform both types of models.

For annotation, it remains to be seen whether different instruction-tuned LLMs can generalize to \textit{any} subjective text classification tasks in different languages, especially if these tasks and languages are not well-represented in the training data. Recent studies have shown the importance of considering human label variation \cite{basile-etal-2021-need, plank-2022-problem}, i.e., the disagreement between human annotators, as a source of information rather than a problem. However, how this human label variation also applies to LLMs remains unexplored. 

\section{Discussion}
Our results indicate that treating LLMs as annotators and aggregating their responses is cost-effective and quick. However, we also find that the overall performance is still well below that of even simple supervised models.

\textbf{Human annotation still has a vital role if we focus on performance}. As LLMs become more capable, this edge might diminish to the point where LLM annotation is equivalent to human annotation. As an aside, although all models are likely to improve across the board, we still expect specialization effects, meaning aggregation approaches will likely stay relevant for the foreseeable future.

But what about bias? Human label variation is not only due to different levels of expertise or diligence \cite{snow-etal-2008-cheap}. It can also vary due to differing opinions, definitions, and biases \cite{shah-etal-2020-predictive}.
Specific tasks are subjective by nature \cite{basile-etal-2019-semeval,rottger-etal-2022-two}, but even seemingly objective tasks like part-of-speech tagging can have different interpretations \cite{plank-etal-2014-linguistically}.
\textbf{The current discussion around the moral and ethical alignment of LLMs \cite{liu-etal-2022-aligning} should make us cautious about using these models as annotators in subjective or sensitive tasks.} Aggregation can overcome the biases of any one particular model, but it cannot safeguard against widespread biases.
Lastly, annotation might be exploratory \citep[the ``descriptive'' paradigm in][]{rottger-etal-2022-two}, where the goal is to find the range of human responses.

However, replacing human annotators with LLMs has ramifications beyond performance and bias issues. 
While crowdsourced annotations can be problematic regarding worker exploitation \cite{fort-etal-2011-last}, they often provide low to moderate-income earners with a way to supplement their living. Replacing this option with LLMs is a real-life example of automation making human jobs obsolete. Conversely, it may mitigate the mental health risks associated with annotating toxic or sensitive content, such as racist content or tasks related to mental disorders. \textbf{Hybrid human and LLM annotation might offer a way forward here}.

\section{Conclusion}
We use zero- and few-shot prompting to compare four current instruction-tuned LLMs as annotators on five subjective tasks in four languages. We find specialization across models and tasks.
We leverage this variance similarly to human label variation by aggregating their predictions into a single label. This approach is, on average, substantially better than any individual model. This suggests that label aggregation consistently enhances performance compared to relying on a single LLM. Despite the rapid development of LLMs enhancing generalization to new tasks, aggregation remains a beneficial strategy.

Our findings suggest that practitioners aiming to label large amounts at minimal cost (both financially and time-wise) can benefit from the outlined aggregation approach.
However, we also find that even the best models cannot compete with ``traditional'' supervised classification approaches.
Furthermore, human annotation allows practitioners to encode a specific view or approach in a prescriptive manner or explore the range of responses descriptively \cite{rottger-etal-2022-two}. Relying on LLMs while alignment and bias still need to be solved \cite{mokander2023auditing} makes this approach unsuitable for sensitive applications.

\section*{Limitations}
\label{sec:limitations}
We were unable to compare to closed LLMs like GPT-4. While they are often state-of-the-art on many tasks, their training and setup change frequently and are, therefore, not replicable. Their pay-by-use nature also makes them less affordable for many practitioners than free open models. We do suspect, though, that including closed or generally better models will not change the overall conclusions of this paper.

\section*{Ethical Considerations}
\label{sec:ethics}

The data we use for AC-gender classification only makes a binary distinction (the Trustpilot website allowed users only to choose from two options). We do not assume this to be representative of gender identities and only use this data to test our hypotheses.

The languages we evaluate all come from the Indo-European branch of languages. The selection was due to data availability and our knowledge of languages. While we do not expect results to systematically differ from other languages, we do note that this is conjecture. 

\section*{Acknowledgements}

This project has received funding from the European Research Council (ERC) under the European Union’s Horizon 2020 research and innovation program (grant agreement No. 949944, INTEGRATOR). The research was made possible in part through an unrestricted Google research gift to explore variance in annotation. Flor Miriam Plaza-del-Arco, Debora Nozza, and Dirk Hovy are members of the MilaNLP group and the Data and Marketing Insights Unit of the Bocconi Institute for Data Science and Analysis. 

\section{Bibliographical References}\label{sec:reference}

\bibliographystyle{lrec-coling2024-natbib}
\bibliography{anthology,languageresource}

\bibliographystylelanguageresource{lrec-coling2024-natbib}
\bibliographylanguageresource{languageresource}

\end{document}

%% file: prompts.tex
\begin{figure}[t]
\centering
\resizebox{0.7\linewidth}{!}{
  \begin{tcolorbox}[colback=gray!10,colframe=black!40]
    \fontsize{11}{12}\selectfont 
    \promptbox{SA}{Is the sentiment of this review ``positive'', ``negative'' or ``neutral''?}{blue}
    \promptbox{AC - Gender}{Is this review written by a ``male'' or a ``female''?}{red}
    \promptbox{AC - Age}{Is this review written by a person ``under 35'', or ``over 45''?}{green}
    \promptbox{TD}{Is this review about topic 1, topic 2, topic 3, topic 4, or topic 5?}{yellow}
    \promptbox{HS}{Is this tweet expressing ``hate speech'' or ``non-hate speech''?}{magenta} 
  \end{tcolorbox}
}
\caption{Instructions used to prompt the instruction-tuned LLMs for each classification task.}
\label{fig:prompts}
\end{figure}

%% file: inter_model_agg_trustpilot.tex
\begin{table}[t]
\small
\centering
\setlength{\tabcolsep}{4pt}
\begin{tabular}{cccccc}
\toprule
\multicolumn{2}{c}{\textbf{Task/Language}} & \textbf{Cohen} & \textbf{Fleiss} & \textbf{Krip.} & \textbf{Raw}  \\
\midrule
    \multirow{3}{*}{SA} 
    & EN & 0.708 & 0.705 & 0.703 & 0.837 \\
    & DE & 0.636 & 0.633 & 0.630 & 0.792 \\
    & FR & 0.665 & 0.662 & 0.660 & 0.809 \\
    \midrule
    \multirow{3}{*}{AC-Gender} 
    & EN & 0.299 & 0.279 & 0.229 & 0.615 \\
    & DE & 0.271 & 0.136 & -0.007 & 0.566 \\
    & FR & 0.236 & 0.227 & 0.179 & 0.596 \\
    \midrule
    \multirow{3}{*}{AC-Age} 
    & EN & 0.101 & 0.044 & -0.154 & 0.428 \\
    & DE & 0.068 & 0.040 & -0.124 & 0.596 \\ 
    & FR & 0.099 & 0.093 & 0.014 & 0.679 \\
    \midrule
    \multirow{3}{*}{TD} 
    & EN & 0.510 & 0.495 & 0.477 & 0.622 \\
    & DE & 0.586 & 0.578 & 0.571 & 0.712 \\ 
    & FR & 0.316 & 0.305 & 0.283 & 0.598 \\
    \midrule
    \multirow{2}{*}{HS} 
    & EN & 0.222 & 0.220 & 0.209 & 0.605 \\
    & ES & 0.155 & 0.099 & -0.019 & 0.629 \\
\bottomrule
\end{tabular}
\caption{Inter-model agreement scores.}\label{tab:agreement_trustpilot}
\end{table}

%% file: competence_trustpilot.tex
\begin{table}[htb!]
\small
\centering
\setlength{\tabcolsep}{2pt}
\begin{tabular}{cccccc}
\toprule
\multicolumn{2}{c}{\textbf{Task/Language}} & \textbf{T0} & \textbf{Flan-T5} & \textbf{Flan-UL2} & \textbf{Tk-Instruct} 
\\
    \midrule
    \multirow{3}{*}{SA} 
    & EN & 0.755 & \textbf{0.958} & 0.856 & 0.803 \\
    & DE & 0.724 & \textbf{0.928} & 0.767 & 0.757 \\
    & FR & 0.759 & \textbf{0.909} & 0.796 & 0.789 \\
    \midrule
    \multirow{3}{*}{AC-Gender} 
    & EN & 0.317 & 0.613 & \textbf{0.719} & 0.445 \\
    & DE & 0.152 & 0.267 & \textbf{0.359} & 0.270 \\
    & FR & 0.094 & \textbf{0.699} & 0.599 & 0.601 \\
    \midrule
    \multirow{3}{*}{AC-Age} 
    & EN & 0.255 & 0.146 & \textbf{0.325} & 0.083 \\
    & DE & 0.295 & 0.039 & \textbf{0.418} & 0.014 \\
    & FR & 0.432 & 0.061 & \textbf{0.608} & 0.017 \\
    \midrule
    \multirow{3}{*}{TD} 
    & EN & 0.388 & 0.737 & \textbf{0.933} & 0.511 \\
    & DE & 0.660 & \textbf{0.793} & 0.735 & 0.712 \\
    & FR & 0.556 & 0.327& \textbf{0.492} & 0.210 
    \\
    \midrule
    \midrule
    \multirow{3}{*}{Mean} 
    & EN & 0.429 & 0.614 & \textbf{0.708} & 0.461 \\
    & DE & 0.458 & 0.507 & \textbf{0.569} & 0.438 \\
    & FR & 0.460 & 0.499 & \textbf{0.624} & 0.404 \\
\bottomrule
\midrule
\multirow{2}{*}{HS} 
    & EN & 0.358 & \textbf{0.919} & 0.327 & 0.402 \\
    & ES & \textbf{0.466} & 0.212 & 0.131 & 0.099 \\
\bottomrule

\end{tabular}
\caption{MACE competence scores of each LLM across tasks and languages on the Trustpilot and HatEval datasets. For non-English languages, we use the multilingual mT0 model.}
\label{tab:tab_competence_trustpilot}
\end{table}

%% file: zs_results.tex
\begin{table*}[ht!]
\small
\centering
\resizebox{\textwidth}{!}{%
\setlength{\tabcolsep}{4pt}
\begin{tabular}{cc|cccc|cc|cc}
\multicolumn{2}{c|}{} & \multicolumn{4}{c|}{\textbf{Models}} & \multicolumn{2}{c|}{\textbf{Baselines}} & \multicolumn{2}{c}{\textbf{Aggregate}}\\
\toprule
\multicolumn{2}{c|}{\textbf{Task/Lang.}} & \textbf{T0} & \textbf{Flan-T5} & \textbf{Flan-UL2} & \textbf{Tk-Instruct} 
& \textbf{Most Freq} &  \textbf{Random} &  \textbf{Majority} & \textbf{MACE} \\
\midrule
\multirow{3}{*}{SA} 
& EN & 0.453$^{\star}$ & 0.532$^{\star}$ & 0.482$^{\star}$ & \textbf{0.553}$^{\star}$ & 0.167 & 0.334 & 0.503$^{\star}$ & 0.514$^{\star}$  
\\
& DE & 0.469$^{\star}$ & 0.495$^{\star}$ & 0.433$^{\star}$ & \textbf{0.517}$^{\star}$ &  0.167 & 0.331 & 0.480$^{\star}$ & 0.484$^{\star}$ 
\\
& FR & 0.460$^{\star}$ & 0.518$^{\star}$ & 0.445$^{\star}$ & \textbf{0.528}$^{\star}$ & 0.167 & 0.337 & 0.486$^{\star}$ & 0.490$^{\star}$ 
\\
\midrule
\multirow{3}{*}{AC-Gender} 
& EN & 0.516 & 0.594$^{\star}$ & \textbf{0.624}$^{\star}$ & 0.541$^{\star}$ & 
0.337 & 0.501 & 0.617$^{\star}$ & 0.623$^{\star}$ 
\\
& DE & 0.456 & 0.437 & 0.447 & 0.431 & 0.334 & \textbf{0.497} & 0.458 & 0.485 
\\
& FR & 0.428 & 0.573$^{\star}$ & 0.566$^{\star}$ & 0.563$^{\star}$ & 0.335 & 0.503 & 0.577$^{\star}$ & \textbf{0.579}$^{\star}$ 
\\
\midrule
\multirow{3}{*}{AC-Age} 
& EN & 0.495 & 0.442 & 0.516$^{\star}$ & 0.397 & 
0.336 & 0.497 & 0.569$^{\star}$ & \textbf{0.572}$^{\star}$ 
\\
& DE & 0.458 & 0.366 & \textbf{0.503} & 0.344 & 0.334 & 0.500 & 0.422 & 0.499
\\
& FR & 0.497 & 0.375 & \textbf{0.550}$^{\star}$ & 0.343 & 
0.335 & 0.500 &  0.443 & 0.542$^{\star}$
\\
\midrule
\multirow{3}{*}{TD} 
& EN & 0.558$^{\star}$ & 0.579$^{\star}$ & 0.588$^{\star}$ & 0.567$^{\star}$ & 0.085 & 0.195 & 0.588$^{\star}$ & \textbf{0.596}$^{\star}$ 
\\
& DE & 0.506$^{\star}$ & 0.514$^{\star}$ & 0.513$^{\star}$ & 0.493$^{\star}$ & 0.105 & 0.193 & 0.516$^{\star}$ & \textbf{0.520}$^{\star}$ \\
& FR & \textbf{0.314}$^{\star}$ & 0.271$^{\star}$ & 0.264$^{\star}$ & 0.257$^{\star}$ & 0.096 & 0.193 & 0.281$^{\star}$ & 0.293$^{\star}$  \\
\midrule
\midrule
\multirow{3}{*}{Mean} 
& EN & 0.506 & 0.537 & 0.553 & 0.515 & 0.231 & 0.382 & 0.569 & \textbf{0.576} \\
& DE & 0.472 & 0.453 & 0.474 & 0.446 & 0.235 & 0.380 & 0.469 & \textbf{0.497} \\
& FR & 0.425 & 0.434 & 0.456 & 0.423 & 0.233 & 0.383 & 0.447 & \textbf{0.476} \\
\bottomrule
\midrule
\multirow{2}{*}{HS} 
& EN & 0.621$^{\star}$ & 0.726$^{\star}$ & 0.670$^{\star}$ & 0.579$^{\star}$ & 
0.367 & 0.490 & 0.717$^{\star}$  & \textbf{0.726}$^{\star}$  
\\
& ES & 0.601$^{\star}$ & 0.532$^{\star}$ & 0.519 & 0.449 & 
0.370 & 0.492 & 0.533$^{\star}$ & \textbf{0.603}$^{\star}$
\\
\bottomrule

\end{tabular}
}
\caption{Zero-shot Macro-F1 results obtained by the LLMs on the Trustpilot and HatEval tasks, the baselines and the aggregation methods. Best result per language and task is shown in bold. Significant improvement over Random baseline ($^{\star}: p \leq 0.01$) with bootstrap sampling. For non-English languages, we use the multilingual mT0 model.} \label{tab:zs_results}
\end{table*}

%% file: results_comp_supervised.tex
\begin{table}[ht!]
\small
\centering
\resizebox{\columnwidth}{!}{%
\setlength{\tabcolsep}{4pt}
\begin{tabular}{cc|c|cc}
&& \textbf{best} & \multicolumn{2}{|c}{\textbf{supervised}}
\\
\multicolumn{2}{c|}{\textbf{Task/Language}} & \textbf{ZSL} & \textbf{Standard ML}   & \textbf{Transformer} \\
\toprule
\multirow{3}{*}{SA} 
& EN & 0.553 & 0.610 & \textbf{0.680}
\\
& DE & 0.517 & 0.610 & \textbf{0.677}
\\
& FR & 0.528 & 0.612 & \textbf{0.706}
\\
\midrule
\multirow{3}{*}{AC-Gender} 
& EN & 0.624 & 0.601 & \textbf{0.638}
\\
& DE & 0.497 & 0.540 & \textbf{0.629}
\\
& FR & 0.579 & 0.546 & \textbf{0.650}
\\
\midrule
\multirow{3}{*}{AC-Age} 
& EN & 0.572 & 0.620 & \textbf{0.636}
\\
& DE & 0.503 & 0.602 & \textbf{0.611}
\\
& FR & 0.550 & 0.540 & \textbf{0.568}
\\
\midrule
\multirow{3}{*}{TD} 
& EN & 0.596 & 0.656 & \textbf{0.705}
\\
& DE & 0.520 & 0.605 & \textbf{0.671}
\\
& FR & 0.314 & 0.385 & \textbf{0.444}
\\
\midrule
\multirow{2}{*}{HS} 
& EN & \textbf{0.726} & 0.451 & 0.416
\\
& ES & 0.603 & 0.701 & \textbf{0.752}
\\
\bottomrule
\end{tabular}
}
\caption{Macro-F1 results for best ZSL model (Table \ref{tab:zs_results}), compared to previous supervised results on the same datasets.} \label{tab:results_trustpilot_comp}
\end{table}